\documentclass[lettersize,journal]{IEEEtran}

\usepackage{amsmath,amsfonts}
\usepackage{algorithmic}
\usepackage{algorithm}
\usepackage{array}
\usepackage[caption=false,font=normalsize,labelfont=sf,textfont=sf]{subfig}
\usepackage{textcomp}
\usepackage{stfloats}
\usepackage{url}
\usepackage{verbatim}
\usepackage{graphicx}
\usepackage{bm}
\usepackage{makecell}
\usepackage{multirow}
\usepackage{color}

\begin{document}

\title{RIFT2: Speeding-up RIFT with A New Rotation-Invariance Technique}

\author{Jiayuan Li, Pengcheng Shi, Qingwu Hu, and Yongjun Zhang
\thanks{This paper was finished and submitted in July 2022.}}

%\author{
%\name{Jiayuan Li$^1$, Pengcheng Shi$^{*,2}$, Qingwu Hu$^1$, and Yongjun Zhang$^{*,1}$}
%\affil{1. School of Remote Sensing and Information Engineering, Wuhan University, Wuhan 430072, China}
%\affil{2. School of Computer Science, Wuhan University, Wuhan 430072, China}
%}

% The paper headers
\markboth{Journal of \LaTeX\ Class Files,~Vol.~14, No.~8, August~2021}%
{Shell \MakeLowercase{\textit{et al.}}: A Sample Article Using IEEEtran.cls for IEEE Journals}

%\IEEEpubid{0000--0000/00\$00.00~\copyright~2021 IEEE}
% Remember, if you use this you must call \IEEEpubidadjcol in the second
% column for its text to clear the IEEEpubid mark.

\maketitle

\begin{abstract}
Multimodal image matching is an important prerequisite for multisource image information fusion. Compared with the traditional matching problem, multimodal feature matching is more challenging due to the severe nonlinear radiation distortion (NRD). Radiation-variation insensitive feature transform (RIFT)~\cite{li2019rift} has shown very good robustness to NRD and become a baseline method in multimodal feature matching. However, the high computational cost for rotation invariance largely limits its usage in practice. In this paper, we propose an improved RIFT method, called RIFT2. We develop a new rotation invariance technique based on dominant index value, which avoids the construction process of convolution sequence ring. Hence, it can speed up the running time and reduce the memory consumption of the original RIFT by almost 3 times in theory. Extensive experiments show that RIFT2 achieves similar matching performance to RIFT while being much faster and having less memory consumption. The source code will be made publicly available in \url{https://github.com/LJY-RS/RIFT2-multimodal-matching-rotation}.
\end{abstract}

%\begin{keywords}
%Multimodal image matching; Feature matching; RIFT; Image fusion; SAR-optical; Infrared-optical.
%\end{keywords}

\section{Introduction}
\IEEEPARstart{I}{mage} matching is a fundamental technique in remote sensing and computer vision, which has been widely used in visual perception,  e.g., photogrammetric digital surface reconstruction, structure-from-motion, simultaneous localization and mapping, virtual reality, target tracking, etc. Traditional image feature matching methods such as scale-invariant feature transform (SIFT)~\cite{lowe2004distinctive}, speeded-up robust features (SURF)~\cite{bay2006surf}, and ORB~\cite{rublee2011orb} are generally designed for same-source matching, which are very sensitive to nonlinear radiation distortion (NRD).

Multimodal images are captured by different types of imaging sensors, which generally display huge apparent differences on the same physical object. For example, optical images and synthetic aperture radar (SAR) images are multimodal. Besides, map-optical, infrared-optical, night-optical, and depth-optical are typical multimodal remote sensing images. These images generally suffer from severe NRD, which becomes the bottleneck issue of multimodal image matching. Recently, multimodal image matching gains more attention and becomes a hotspot in feature matching. It mainly contains two types of methods~\cite{li2017robust,li2019rift}, i.e., area-based matching~\cite{viola1997alignment,ye2016hopc,ye2017robust} and feature matching~\cite{li2019rift,li2022lnift,yao2022multi}.

Area-based matching is also known as template matching or patch matching, which seeks the patch in the reference image that has the highest similarity with the template~\cite{li2017robust2}. Histogram of orientated phase congruency (HOPC)~\cite{ye2017robust} introduces an extended phase congruency model that contains both magnitude and orientation information. Channel features of orientated gradients (CFOG)~\cite{ye2019fast} provides a pixel-wise feature representation and a three-dimensional fast Fourier transform (FFT) matching measure. Recently, deep learning show great potential for patch matching. For example, Hughes et al.~\cite{hughes2018mining} proposed a hard-negative mining technique for SAR-optical registration. Fang et al.~\cite{fang2021sar} combined Siamese U-net and FFT to tackle significant heterogeneous characteristics. Since area-based methods only perform a two-dimensional translation search, they are sensitive to complex geometric transformations such as rotation and perspective variances. Hence, HOPC and CFOG require prior geographic information for initializations.

Feature matching methods are more flexible and robust to geometric distortions compared with area-based ones. They generally contain three stages, including keypoint detection, local feature description, and one-to-one matching. Ma et al.~\cite{ma2016remote} provided a new definition for the gradient to overcome the intensity difference between multisensor images and proposed a SIFT-like method called position-scale-orientation SIFT (PSO-SIFT). Xiang et al.~\cite{xiang2018sift} improved the SIFT to be suitable for optical-SAR matching, named optical-SAR SIFT (OS-SIFT), which constructs Harris scale spaces based on the ratio of exponentially weighted averages (ROEWA) and Sobel operators. Similar to OS-SIFT, Yu et al.~\cite{yu2021universal} also used ROEWA and Sobel operators for Harris scale space construction and proposed a SIFT variant to describe features. Hughes et al.~\cite{hughes2020deep} presented a deep learning framework that consists of three convolutional neural networks to predict matchable regions, generate correspondence heatmap, and remove outliers via binary classification. However, these methods are only suitable for specific image types. The first feature matching method that is suitable for different types of multimodal images may be the radiation-variation insensitive feature transform (RIFT)~\cite{li2019rift}. It proposes a maximum index map instead of the gradient for feature description and uses a convolution sequence ring to achieve rotation invariance. Locally normalized image feature transform (LNIFT)~\cite{li2022lnift} proposes a local normalization filter to convert two different modalities into the same intermediate modal in the spatial domain. Among these methods, RIFT almost has become a baseline method in multimodal feature matching. However, its computational complexity is large due to the construction of the convolution sequence ring, which largely limits its usage in practice.

In this paper, we focus on addressing the bottleneck in the RIFT method and proposed an improved variant called RIFT2. We develop a new rotation invariance technique based on the dominant index in the maximum index map (MIM) instead of constructing a convolution sequence ring. With the convolution sequence ring in RIFT, each keypoint in the reference image becomes six features with the same location coordinates but different MIMs. As a result, the number of total features of the reference image increases by six times (The number of features of target images remains the same). Therefore, our new dominant index technique can speed up the running time and reduce the memory consumption of the original RIFT by almost 3 times in theory. Extensive experiments demonstrate that replacing the old enumeration strategy with the proposed new technique almost does not cause performance degradation, but largely improves computational efficiency and reduces memory consumption.

\begin{figure}[!t]
\centering
\includegraphics[width=8.8cm]{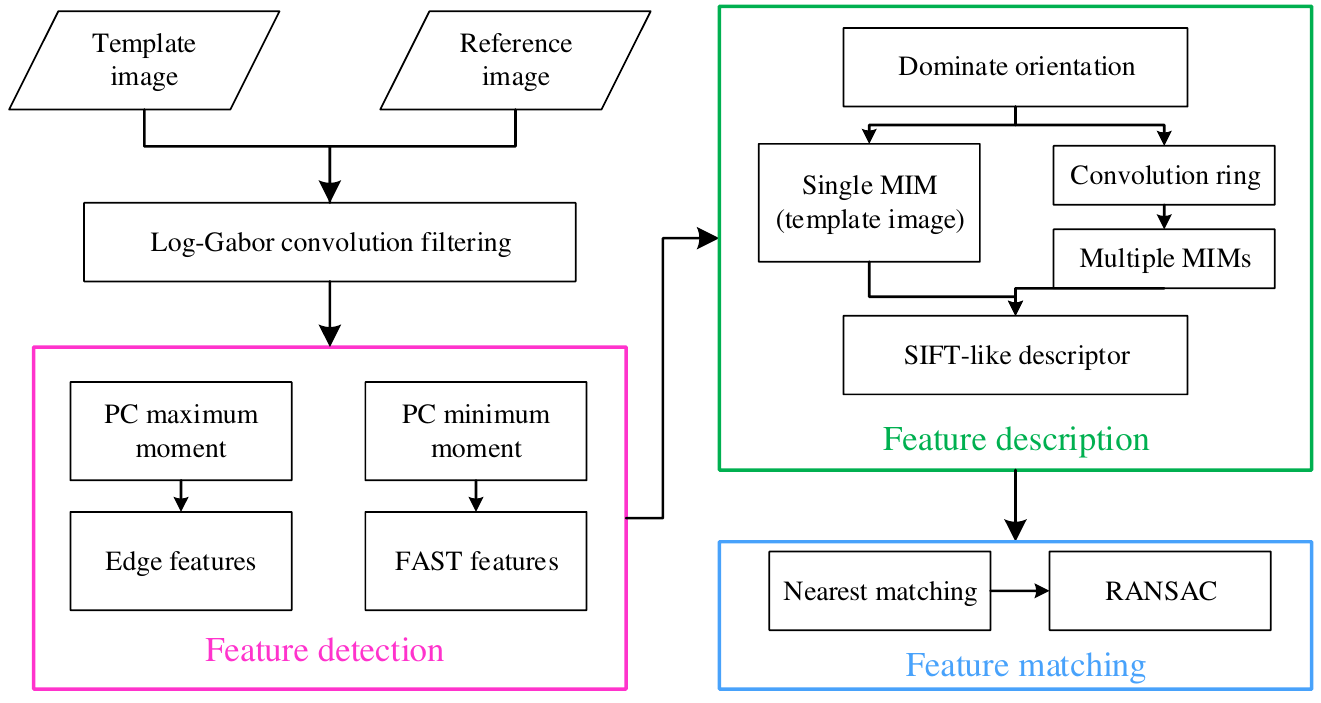}
\caption{The framework of RIFT, which contains detection, description, and matching.}
\label{fig1}
\end{figure}

\section{RIFT Revisit}

Figure \ref{fig1} is the workflow of RIFT, which also contains three main stages, i.e., detection, description, and matching. RIFT detects keypoints on the phase congruency (PC) layer and describes features based on a maximum index map (MIM). It achieves rotation invariance by a convolution sequence ring. Then, RANSAC-like methods~\cite{fischler1981random,9007014} are used to filter outliers.

\subsection{Feature Detection}

To obtain a large number of distinctive feature points stably in different modal images, RIFT uses the phase congruency measure for keypoint detection. Specifically, it first calculates a one-dimensional phase congruency for each orientation of the log-Gabor convolutions and analyzes the change of moment with orientations; then, RIFT extracts corner points on the minimum moment image using the FAST detector since the minimum moment is equivalent to cornerness and detects edge features on the maximum moment map that represents edge information. Corner points have high repeatability, but the number of kepoints is small. Edge features are complementary to corner ones. Thus, RIFT integrates both corner features and edge features for image matching, which can not only ensure high repeatability but also ensure that the number of features is high enough. This lays the foundation for high matching accuracy and a large number of correct matches.

\subsection{Feature Description}\label{FD}

Both intensity and gradient have only linear radiation invariance and are very sensitive to NRD. Thus, RIFT proposes a new concept called the maximum index map (MIM) for feature description based on the Log-Gabor convolution sequence. Specifically, it first computes the sum of amplitudes ${A_o}(\bm{x})$ along each orientation $o$ based on the response components of the Log-Gabor filter and arranges these $N_o$ amplitudes in order to obtain a multi-channel convolution map $\{ A_o^\omega (\bm{x})\} _1^{{N_o}}$, where $N_o$ is the number of orientations and $\omega  = 1,2,...,{N_o}$. Then, for each pixel position $(\bm{x}_i)$, RIFT finds the maximum value ${A_{\max }}(\bm{x}_i)$ and its channel index ${\omega _{\max }}$, i.e., $[{A_{\max }}(\bm{x}_i),{\omega _{\max }}] = \max \{ \{ A_o^\omega (\bm{x}_i)\} _1^{{N_o}}\} $. These indexes correspond to maximum values from the MIM. Finally, a SIFT-like histogram technique is used for feature vector encoding. For each keypoint, its local MIM patch is divided into $6\times 6$ grids. A $N_o$-bins histogram is calculated for each grid and all these histograms are concatenated to obtain a $6\times 6\times N_o$ dimension RIFT feature vector.

\subsection{Rotation Invariance}

As analyzed in RIFT, the construction of MIM is highly related to orientations. To eliminate the effect of rotations, RIFT introduces a convolution sequence ring structure to generate multiple MIMs for each feature in the reference image. Specifically, RIFT turns the Log-Gabor sequence into an end-to-end ring structure. For each feature in the template image, RIFT constructs a single MIM $MIM_t$; for each feature in the reference image, it generates $N_o=6$ MIMs $\{ MIM_r^\omega\} _1^{{N_o}}$ with different initial layers. Therefore, the $N_r$ keypoints in the reference image generate $6N_r$ description vectors, which increase the computational complexity of reference image feature description by 6 times. Moreover, the $N_t \times N_r$ ($N_t$ is the number of keypoints in the target image) matching becomes a $6 \times N_t \times N_r$ matching problem, witch also increases the running time of the matching stage by 6 times.

\begin{figure}[!t]
\centering
\includegraphics[width=8.8cm]{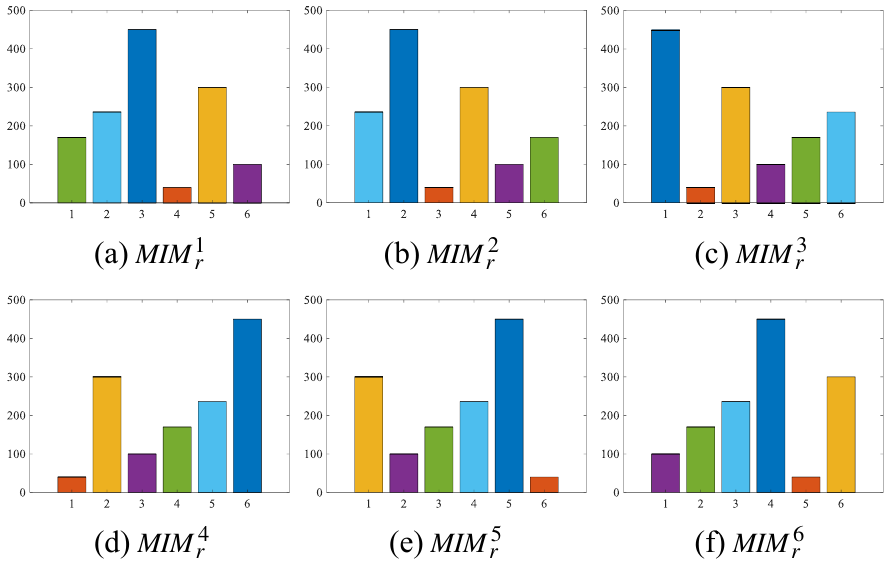}
\caption{The relationship between histogram and different MIMs $\{ MIM_r^\omega\} _1^{{6}}$. }
\label{fig2}
\end{figure}

\section{Dominate Index Value}

The bottleneck of RIFT lies in the construction of multiple MIMs. Since these MIMs are obtained by transforming the initial layer of the Log-Gabor convolution ring, the change of MIM must satisfy a certain law. Thus, we first analyze the relationship among these MIMs. Specifically, given an image patch with $36\times 36$ pixels, we generate 6 MIMs $\{ MIM_r^\omega\} _1^{{6}}$ according to Section \ref{FD}. $MIM_r^\omega$ represents that the $\omega$-th layer is the initial layer. Then, the index of $\omega$-th layer of the Log-Gabor sequence becomes 1 in the $MIM_r^\omega$. If we change the initial layer to be the second layer of Log-Gabor sequence, then, $MIM_r^1(\bm{x}_i)=2$ becomes $MIM_r^2(\bm{x}_i)=1$ and $MIM_r^1(\bm{x}_i)=4$ becomes $MIM_r^2(\bm{x}_i)=3$. Suppose the histogram of $MIM_r^1$ is $\left\{ {\left( {1,170} \right),\left( {2,236} \right),\left( {3,450} \right),\left( {4,40} \right),\left( {5,300} \right),\left( {6,100} \right)} \right\}$. It becomes $\left\{ \!{\left( {1,236} \right),\!\left( {2,450} \right),\!\left( {3,40} \right),\!\left( {4,300} \right),\!\left( {5,100} \right),\!\left( {6,170} \right)} \right\}$ in $MIM_r^2$, as shown in Fig. \ref{fig2}. Therefore, the histogram bins shift sequentially when the MIM is changed from $MIM_r^\omega$ to $MIM_r^j$. If we find the peak of the histogram, we can set the index of the peak histogram bin to be 1 and recode other histogram bins. Then, the recoded histograms of different MIMs are the same, which are independent of rotations.

\begin{figure}[!t]
\centering
\includegraphics[width=8.8cm]{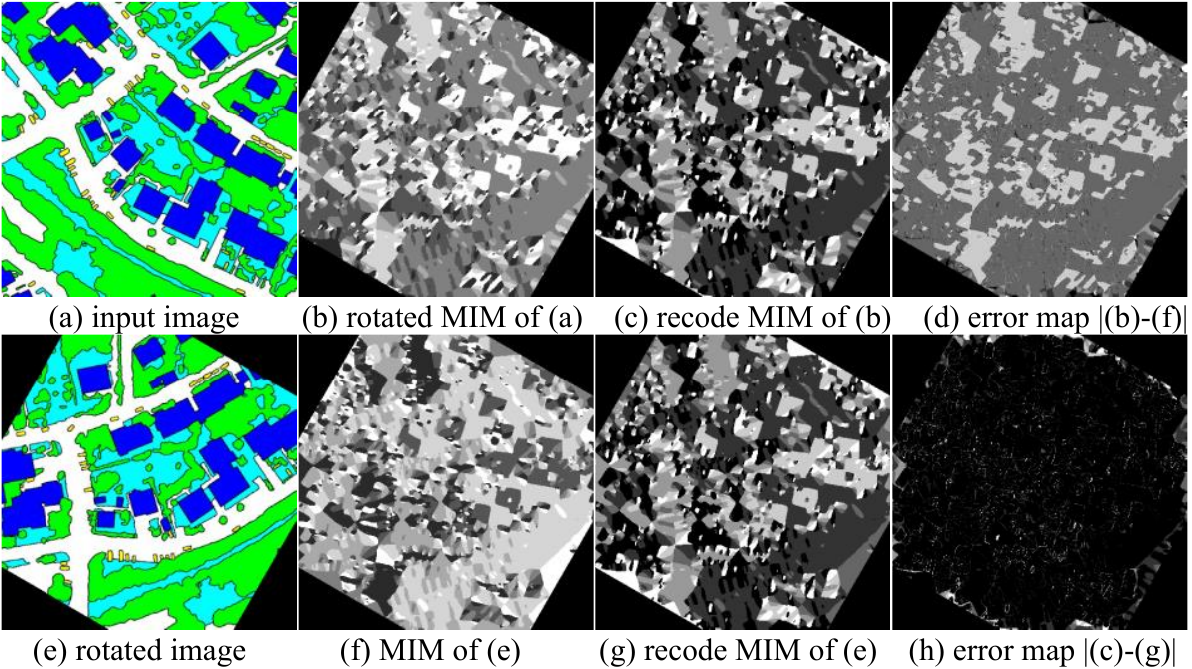}
\caption{An example to compare the original MIM and our recoded MIM. }
\label{fig3}
\end{figure}

\begin{table*}[t]
\renewcommand{\arraystretch}{1.5}
\centering
\footnotesize
 \setlength{\tabcolsep}{8mm}
\caption{Detailed settings of compared algorithms}
\begin{tabular}{|c|c|c|}\hline
{Method} & {Main parameters} & {Implementations}\\ \hline\hline
\rule{0pt}{14pt}
SIFT & \makecell[c]{Keypoint number: 5000; \\patch size: 96; contrast threshold: 0.001} & \makecell[c]{C++ code:\\
\begin{footnotesize}
{https://www.vlfeat.org/overview/sift.html}
\end{footnotesize}}\\
\rule{0pt}{14pt}
PSO-SIFT & \makecell[c]{Keypoint number: 5000; \\patch size: 96; contrast threshold: 0.001}& \makecell[c]{MATLAB code:\\
\begin{footnotesize}
{https://github.com/ZeLianWen/Image-Registration}
\end{footnotesize}}\\
\rule{0pt}{14pt}
OS-SIFT & \makecell[c]{Keypoint number: 5000; \\patch size: 96; Harris threshold: 0.001} & \makecell[c]{MATLAB code:\\
\begin{footnotesize}
{https://sites.google.com/view/yumingxiang/}
\end{footnotesize}}\\
\rule{0pt}{14pt}
RIFT & \makecell[c]{Keypoint number: 5000; \\patch size: 96; FAST threshold: 0.001} & \makecell[c]{MATLAB code:\\
\begin{footnotesize}
{https://ljy-rs.github.io/web/}
\end{footnotesize}}\\
\rule{0pt}{14pt}
Our RIFT2 & \makecell[c]{Keypoint number: 5000; patch size: 96; \\FAST threshold: 0.001; dominate ratio: 0.8}
& \makecell[c]{C++ code:\\
\begin{footnotesize}
{https://ljy-rs.github.io/web/}
\end{footnotesize}}\\
\hline
\end{tabular}
  \label{table1}
\end{table*}

Based on the above analysis, the Log-Gabor convolution ring and multiple MIMs construction stages can be avoided. In detail, we first compute a single MIM for each keypoint from either the template image or the reference image. Then, we calculate the histogram of the MIM patch around the feature and find the largest value $v$ and the dominate index $s$,
\begin{equation}
\label{eq1}
[v, s] = argmax(hist(MIM))
\end{equation}
Inspired by the dominate orientation in SIFT, dominate index $s$ can be used to normalize the histograms,
\begin{equation}
\label{eq2}
MI{M_n}({\bm{x}_i}) = \left\{ {\begin{array}{*{20}{c}}
{MIM({\bm{x}_i}) - s + 1}&{MIM({\bm{x}_i}) \ge s}\\
{MIM({\bm{x}_i}) + {N_o} - s + 1}&{otherwise}
\end{array}} \right.
\end{equation}
where $MIM_n$ is the recoded MIM layer, which is rotation invariant. To improve the stability of matching, we create two MIMs for cases where the dominate index is not very distinctive. Namely, if the second highest bin is within 80\% of the peak, this bin is also used to create a new MIM.

Fig. \ref{fig3} provides an example to compare the MIM of RIFT and our recoded MIM. Fig. \ref{fig3}(a) and Fig. \ref{fig3}(e) are a label map image and its rotated version, respectively. Fig. \ref{fig3}(b) and Fig. \ref{fig3}(f) are the MIMs of Fig. \ref{fig3}(a) and Fig. \ref{fig3}(d), respectively, where Fig. \ref{fig3}(b) is rotated so that the rotation difference between Fig. \ref{fig3}(b) and Fig. \ref{fig3}(f) is eliminated. Fig. \ref{fig3}(c) and Fig. \ref{fig3}(g) are our recoded MIMs of Fig. \ref{fig3}(b) and Fig. \ref{fig3}(f), respectively. Fig. \ref{fig3}(d) is the error map between Fig. \ref{fig3}(b) and Fig. \ref{fig3}(e). Fig. \ref{fig3}(h) is the error map between Fig. \ref{fig3}(c) and Fig. \ref{fig3}(g). As can be seen, the original MIM is sensitive to rotations while our recoded MIM gets zero error in most pixels. Namely, the values of our new MIMs are invariant to rotations. Thus, a single MIM is enough for describing an image, which avoids the construction of the Log-Gabor ring and multiple MIMs.

\begin{figure*}[!h]
\centering
\includegraphics[width=18cm]{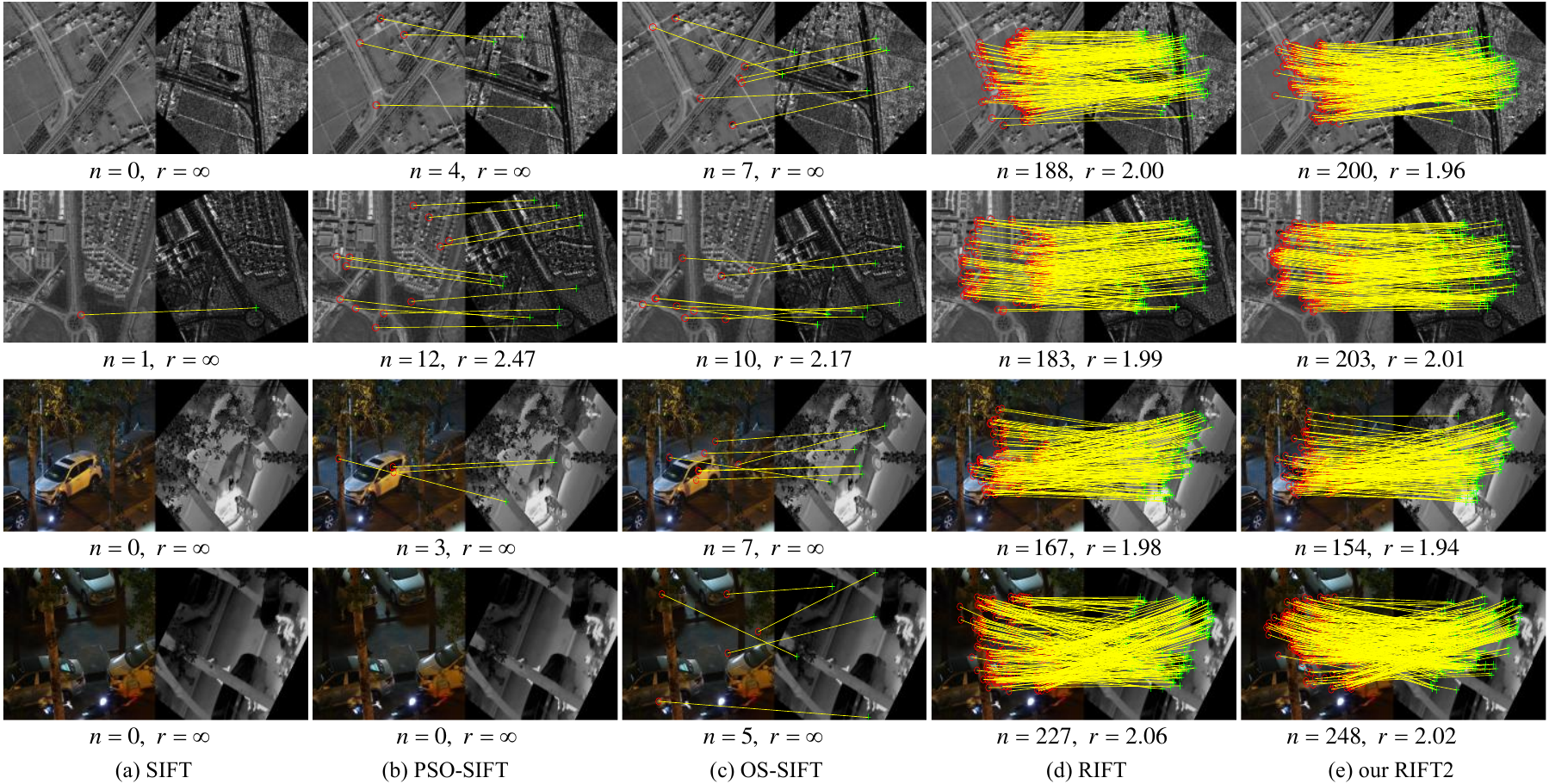}
\caption{{Qualitative comparison results. The first two raws display the results on the optical-SAR dataset and the last two raws are the results on the optical-infrared dataset. Keypoints are shown as red circles and green crosshairs; correct matches are represented as yellow lines. If the matching is fail, its RMSE is denoted by $r=\infty $.}}
\label{fig4}
\end{figure*}

\section{Experiments}

Here, we qualitatively and quantitatively evaluated the proposed RIFT2 on two multimodal datasets, i.e., SAR-optical and thermal infrared-optical. The SAR-optical dataset consists of 100 image pairs that are randomly picked from the Dataset 2 of LNIFT, and the infrared-optical dataset with also 100 image pairs is randomly selected from the Dataset 4. Both the datasets suffer from rotation changes. Each image pair of both two datasets is assigned a ground truth transformation with a rotation matrix and a translation vector. We compare RIFT2 with SIFT~\cite{lowe2004distinctive}, PSO-SIFT~\cite{ma2016remote}, OS-SIFT~\cite{xiang2018sift}, and RIFT~\cite{li2019rift}. For a fair comparison, we fix the maximum number of keypoints to 5000 and set the patch size to $96\times 96$ pixels for all methods. Moreover, we disable the scale space construction stage of SIFT, PSO-SIFT, and OS-SIFT, since RIFT and RIFT2 are not scale-invariant and the datasets do not suffer from scale changes. We use brute force matching with the nearest feature distance for all compared algorithms. Table \ref{table1} provides the setting details of compared methods.

Three widely used metrics are applied for quantitative evaluations, including correct match number $n$, root mean square error (RMSE) $r$, and success rate $\gamma$. To better evaluate local descriptors, we do not apply RANSAC-like methods~\cite{fischler1981random} or local geometric constraints~\cite{9373914,li2019lam} to detect correct matches for all methods, since RANSAC-like methods may also filter out true inliers. Instead, we use ground truth transformations to extract correct matches, i.e., matches with residuals less than 3 pixels. The same as LNIFT, an image pair is correctly matched if $n \geq 10$. The success rate $\gamma$ represents the number of correctly matched pairs in 100 image pairs. The maximum RMSE is set to 20 pixels. All the experiments are performed on a laptop with an i7-8550U @ 1.8GHz CPU, and 8 GB of RAM.

\subsection{Qualitative Evaluations}

We pick two pairs from each dataset for qualitative comparisons. The SAR images of the first two pairs contain severe speckle noise. The last two image pairs suffer from large lighting variations, i.e, night-time RGB and infrared images. These make the matching quite difficult. The results are displayed in Figure \ref{fig4}. As shown, SIFT completely fails on these pairs, since it only considers linear radiation distortion. Both PSO-SIFT and OS-SIFT perform better than SIFT since they redefined the gradient of SIFT so that it can better adapt to multimodal images. However, their performance is far from enough. They only successfully match on one image pair, i.e., the success rate is only $25\%$. In addition, the number of correct matches is much smaller than RIFT and our RIFT2 even though the matching is successful, and the RMSEs are larger. Both RIFT and our RIFT2 achieve very good performance on all the image pairs. This is expected since the only difference between RIFT and our RIFT2 is the rotation invariance technique. RIFT introduces phase congruency for feature extraction and MIM for description, which is robust to NRD and becomes a baseline for multimodal feature matching. Our new technique is efficient and does not lose the performance of RIFT.

\begin{table*}[!t]
\renewcommand{\arraystretch}{1.5}
  \centering
  \small
  \caption{Quantitative evaluation results. Each value is the average result.}
  \setlength{\tabcolsep}{4.5mm}
{
  \begin{tabular}{|c|c|ccccc|}\hline
 \multirow{2}{*}{Data}& \multirow{2}{*}{Metric} & \multicolumn{5}{c|}{Method}\\\cline{3-7}
  & & SIFT &PSO-SIFT &OS-SIFT&RIFT&Our RIFT2\\\hline\hline
 \multirow{3}{*}{\makecell[c]{SAR-optical}}& RMSE $r$ (pixels)$\downarrow$& 19.64 & 15.31& 13.21& 3.12& {2.79}\\
  & success rate $\gamma$ (\%)$\uparrow$& 2 & 26& 38& 94& {96}\\
  & correct match number $n$ $\uparrow$& 2 & 7& 10& {106}& {102}\\
  & running time $n$ $\downarrow$& 0.87 & 1.04& 1.78& 29.24& {10.93}\\
\hline
 \multirow{3}{*}{\makecell[c]{infrared-optical}}& RMSE $r$ (pixels)$\downarrow$& 19.45 & 14.21& 17.13& 2.45& {2.62}\\
 & success rate $\gamma$ (\%)$\uparrow$& 3 & 32& 16& 98& 97\\
 & correct match number $n$ $\uparrow$& 2 & 8& 10& 123& {118}\\
  &  running time $n$ $\downarrow$& 1.75 & 1.92& 2.27& 35.82& {13.24}\\
\hline
\end{tabular}}
\label{table2}
\end{table*}

\subsection{Quantitative Evaluations}

Table \ref{table2} reports the quantitative results, including RMSE (lower is better), success rate (higher is better), number of correct matches (higher is better), and running time (lower is better). As can be seen, SIFT fails on most pairs, whose success rate is lower than 5\%. PSO-SIFT and OS-SIFT get success rates of 20\% $\sim$ 30\%, which is much higher than SIFT. PSO-SIFT performs better than OS-SIFT on the optical-infrared dataset, while OS-SIFT is better than PSO-SIFT on the optical-SAR dataset. The reason is that OS-SIFT is specially designed for optical-SAR matching. It uses the ratio of exponentially weighted averages for SAR image scale space construction. However, this ratio may get very poor performance on other types of images. Both RIFT and our RIFT2 achieve impressive results on these two datasets. Their success rates are around 96\% and the numbers of correct matches are higher than 100.

The average success rates of these five compared methods on the two datasets are 2.5\%, 29\%, 27\%, 96\%, and 96.5\%, respectively. RIFT and RIFT2 almost gain a growth rate of more than 65\% compared with PSO-SIFT and OS-SIFT. In terms of the number of correct matches $n$, the results of SIFT, PSO-SIFT, OS-SIFT, RIFT, and our RIFT2 are 2, 7.5, 10, 114.5, and 110, respectively. The correct matches of RIFT or RIFT2 are 10 times of the ones of PSO-SIFT and OS-SIFT. Our RMSE is slightly better than RIFT, i.e., smaller than 3 pixels. For feature matching methods, the success rate is relatively more important than RMSE, since feature-based methods generally can not achieve very high feature location accuracy due to the NRD or speckle noise. Hence, for applications that require very high geometric precision, template-based methods and sub-pixel accuracy strategies are often used to further refine the results of feature-based methods.

In terms of running time, SIFT, PSO-SIFT, and OS-SIFT are much faster than RIFT and RIFT2. The reason is that they generally detect much fewer keypoints than RIFT, which largely reduce computational complexity. Although the keypoint numbers of RIFT and RIFT2 are the same, our RIFT2 reduces the running time of RIFT by almost 3 times.  The reason is that our method does not construct the Log-Gabor convolution ring and multiple MIMs. Hence, it reduces the number of descriptors of the reference image by 6 times. Note that the descriptors of the target image remain the same.

\section{Conclusions}

This paper proposed an improved RIFT algorithm called RIFT2 that introduces a new rotation invariance technique to address the bottleneck in the RIFT. It uses the dominant index via histogram analysis instead of constructing a convolution sequence ring for rotation invariance. With this technique, multiple MIMs will no longer be required for the reference image. Hence, the computational complexity and memory
footprint can be largely reduced. Extensive experiments show that our proposed new technique can almost speeding-up the original RIFT by 3 times without causing performance degradation. Our future work will focus on scale invariance (scale space construction) and near real-time performance (GPU implementation).

\section*{Acknowledgments}

This work was supported by National Natural Science Foundation of China (No. 42030102 and 41901398)

\bibliographystyle{IEEEtran}
\bibliography{reference.bib}

\end{document}